\documentclass[10pt,twocolumn,letterpaper]{article}

\usepackage[pagenumbers]{cvpr} 
\definecolor{cvprblue}{rgb}{0.21,0.49,0.74}
\usepackage[pagebackref,breaklinks,colorlinks,allcolors=cvprblue]{hyperref}

\def\paperID{*****} 
\def\confName{CVPR}
\def\confYear{2025}

\title{Skeleton-Guided Spatial-Temporal Feature Learning for Video-Based Visible-Infrared Person Re-Identification}

\author{Wenjia Jiang \quad Xiaoke Zhu \quad Jiakang Gao \quad Di Liao\\
Henan University, Kaifeng 475001, China\\
{\tt\small jwj1342@henu.edu.cn, xkzhu@henu.edu.cn, gjk@henu.edu.cn, liaodi@henu.edu.cn}
}
\usepackage{tabularray}
\usepackage{pifont}

\begin{document}
\maketitle
\begin{abstract}
Video-based visible-infrared person re-identification (VVI-ReID) is challenging due to significant modality feature discrepancies. Spatial-temporal information in videos is crucial, but the accuracy of spatial-temporal information is often influenced by issues like low quality and occlusions in videos. Existing methods mainly focus on reducing modality differences, but pay limited attention to improving spatial-temporal features, particularly for infrared videos.
To address this, we propose a novel Skeleton-guided spatial-Temporal feAture leaRning \textbf{(STAR)} method for VVI-ReID. By using skeleton information, which is robust to issues such as poor image quality and occlusions, STAR improves the accuracy of spatial-temporal features in videos of both modalities.
Specifically, STAR employs two levels of skeleton-guided strategies: frame level and sequence level. At the \textbf{frame level}, the robust structured skeleton information is used to refine the visual features of individual frames. At the \textbf{sequence level}, we design a feature aggregation mechanism based on skeleton key points graph, which learns the contribution of different body parts to spatial-temporal features, further enhancing the accuracy of global features.
Experiments on benchmark datasets demonstrate that STAR outperforms state-of-the-art methods. Code will be open source soon.
\end{abstract}    
\section{Introduction}
\label{sec:intro}

\begin{figure}[t]
\centering
\includegraphics[width=1\columnwidth]{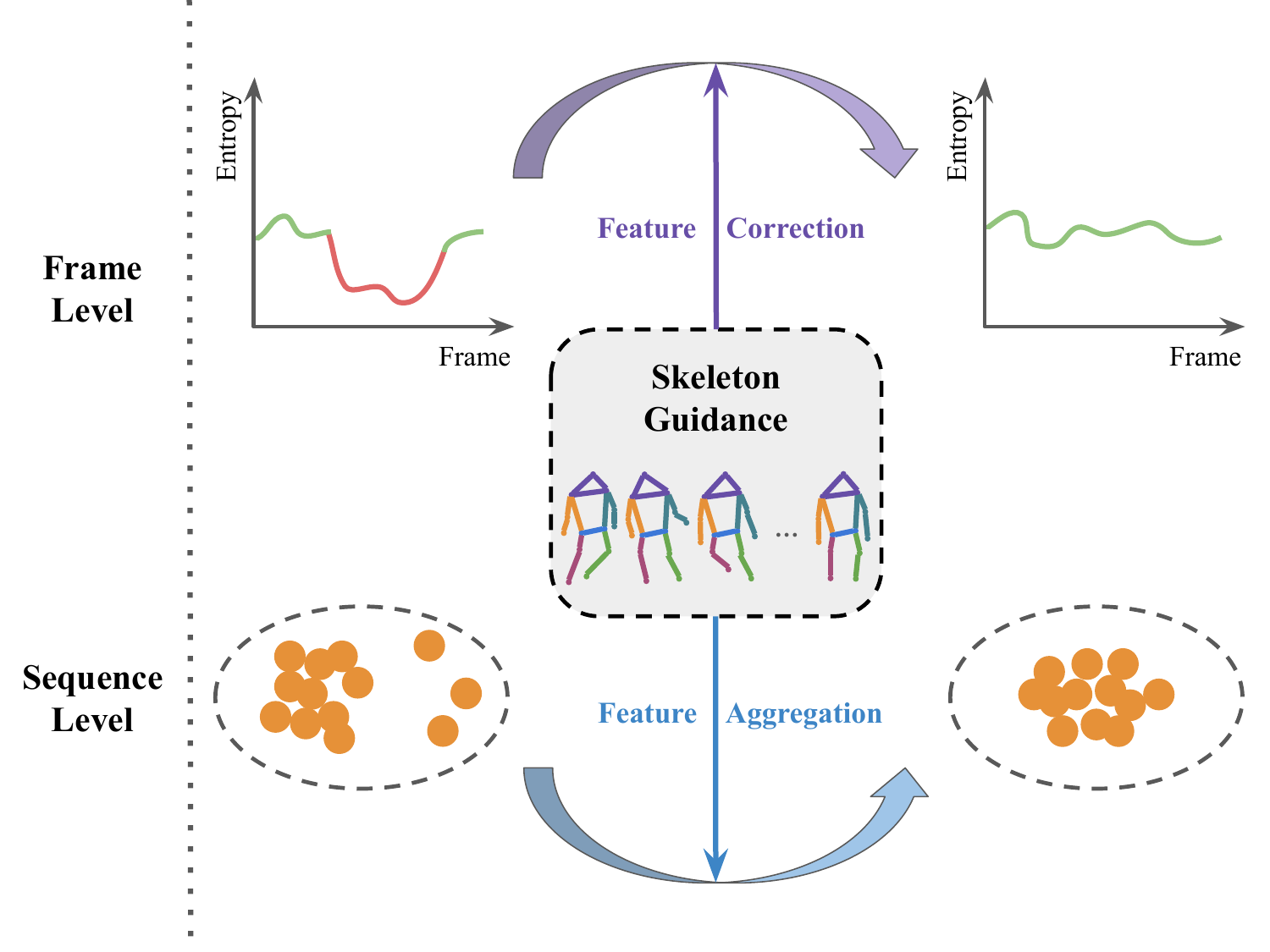} 
\caption{The insights of our method. The Skeleton-guided spatial-Temporal feAture leaRning (\textbf{STAR}) method operates at two levels: At the frame level, skeleton guidance corrects incomplete vision feature information. At the sequence level, it ensures global feature consistency by leveraging skeleton-based guidance.}
\label{motivation}
\end{figure}
Video-based Visible-Infrared Person Re-Identification (VVI-ReID)\cite{VVI-ReID1, VVI-ReID2, MITML, IBAN} is gaining increasing attention due to the rich spatial-temporal information in video sequences. VVI-ReID aims to match the identity of pedestrians captured in video sequences from both visible and infrared cameras. This task presents two primary challenges\cite{primary_challenge, SGIEL}: how to mitigate cross-modality discrepancies and how to exploit the spatial-temporal information in videos.

To address these challenges, researchers have explored various methods, achieving remarkable advancements. To reduce the modality gap, many studies focus on acquiring and utilizing modality-invariant features \cite{lu2020cross,zhang2021learning,hu2022adversarial}. For instance, \cite{du2023enhanced} explores the potential of using a cross-modality shared-specific feature transfer algorithm to mine modality-invariant characteristics. For effective extraction and utilization of spatial-temporal information, methods based on temporal aggregation \cite{li2019multi,li2020multi} and complementary information mining \cite{li2019global,basaran2020efficient} have gained attention.
Temporal aggregation approaches typically treat each video segment as a directed sequence, employing RNN-based or 3D-CNN modules to aggregate features across all frames, thereby extracting temporal features. Complementary information mining aims to extract discriminative features by exploring the relationships between unordered frames. Although previous researchers have made progress in addressing challenges in VVI-ReID, real-world issues such as occlusions, viewpoint variations, and low-quality images will impair the effectiveness of frame-level feature extraction. These problems also impair the accuracy of spatial-temporal features at the sequence-level.

Recent studies have shown that skeleton data, recognized for its exceptional anti-interference capabilities and rich spatial-temporal features \cite{shi2019skeleton,xu2021scene,wu2023spatiotemporal}, plays a significant role in many vision-based tasks, such as action recognition \cite{li2017skeleton,li2019actional,Cheng_2020_CVPR,Duan_2022_CVPR} and image-based single-modality person re-identification \cite{10.1145/3538490,zhu2023ecreid,han20223d}. 

Motivated by these works, we intend to introduce the pedestrian skeleton data into the VVI-ReID domain. Our goal is to leverage its strengths to overcome the challenges associated with spatial-temporal features. Intuitively, the interference-resistant properties of skeleton data can ensure effective representation even for partially occluded pedestrians, thereby enhancing the robustness of frame-level image features, while the spatial-temporal information within the skeleton can further enrich the spatial-temporal features obtained at the sequence-level.

Based on these insights, we propose a Skeleton-guided spatial-Temporal feAture leaRning (\textbf{STAR}) approach for VVI-ReID, which can extract more accurate spatial-temporal features through frame-level feature correction and sequence-level feature aggregation. Specifically, at the frame level, the robust structured skeleton features are used to enhance the representation of low-quality frames in the infrared modality, and mitigate local feature distortions caused by visual noise. At the sequence level, we introduce a body part contribution aware skeleton guidance strategy to further improve the effectiveness of the spatial-temporal features. 

The contributions can be summarized as follows:
\begin{itemize}
    \item This is the first work to incorporate skeleton information into the field of VVI-ReID, and propose a Skeleton-guided spatial-Temporal feAture leaRning (STAR) approach to learn more effective spatial-temporal features.
    \item We design a frame-level feature correction module, which exploits the robust structured skeleton information to extract robust features for each frame.
    \item We design a sequence-level feature aggregation module, which further enhance the effectiveness of the spatial-temporal features by learning body part contribution from skeleton data.
    \item Extensive experimental results demonstrate that our method outperforms the state-of-the-art methods on benchmark datasets.
\end{itemize}
\section{Related works}
\label{sec:relatework}
\subsection{Video-based Visible-Infrared Person ReID}

Due to its richer spatial-temporal information compared to image-based cross-modal person re-identification, video-based visible-infrared person re-identification (VVI-ReID) has garnered significant attention. The two primary challenges in this domain include mitigating modality differences and extracting spatial-temporal information. In recent years, researchers have focused on alleviating the modality gap between infrared and visible light in video sequences. \cite{lin2022learning} were the first to introduce the video-based visible-infrared person re-identification dataset HITSZ-VCM and proposed an adversarial learning approach to preserve modality-invariant features. However, this approach led to the loss of valuable modality-independent features. To address this issue, \cite{li2023intermediary} proposed the IBAN model, which uses anaglyph data of pedestrian images as an intermediary to learn modality-independent features and introduced a bidirectional spatial-temporal aggregation module to leverage the spatial-temporal information in video data. Nevertheless, IBAN struggles to effectively capture global spatial features and long-range temporal dependencies in ultra-long sequences. To overcome this limitation, \cite{feng2024cross} proposed a Cross-Modal Spatial-temporal Transformer (CST), which establishes global spatial-temporal relationships by modeling the long-range temporal dependencies of pedestrian features. These studies have made significant advancements, driving progress in the VVI-ReID field. In contrast to existing studies, we propose a different approach, which uses skeletal data as a robust guide for the feature extraction of person videos. 

\begin{figure*}[t]
\centering
\includegraphics[width=1\textwidth]{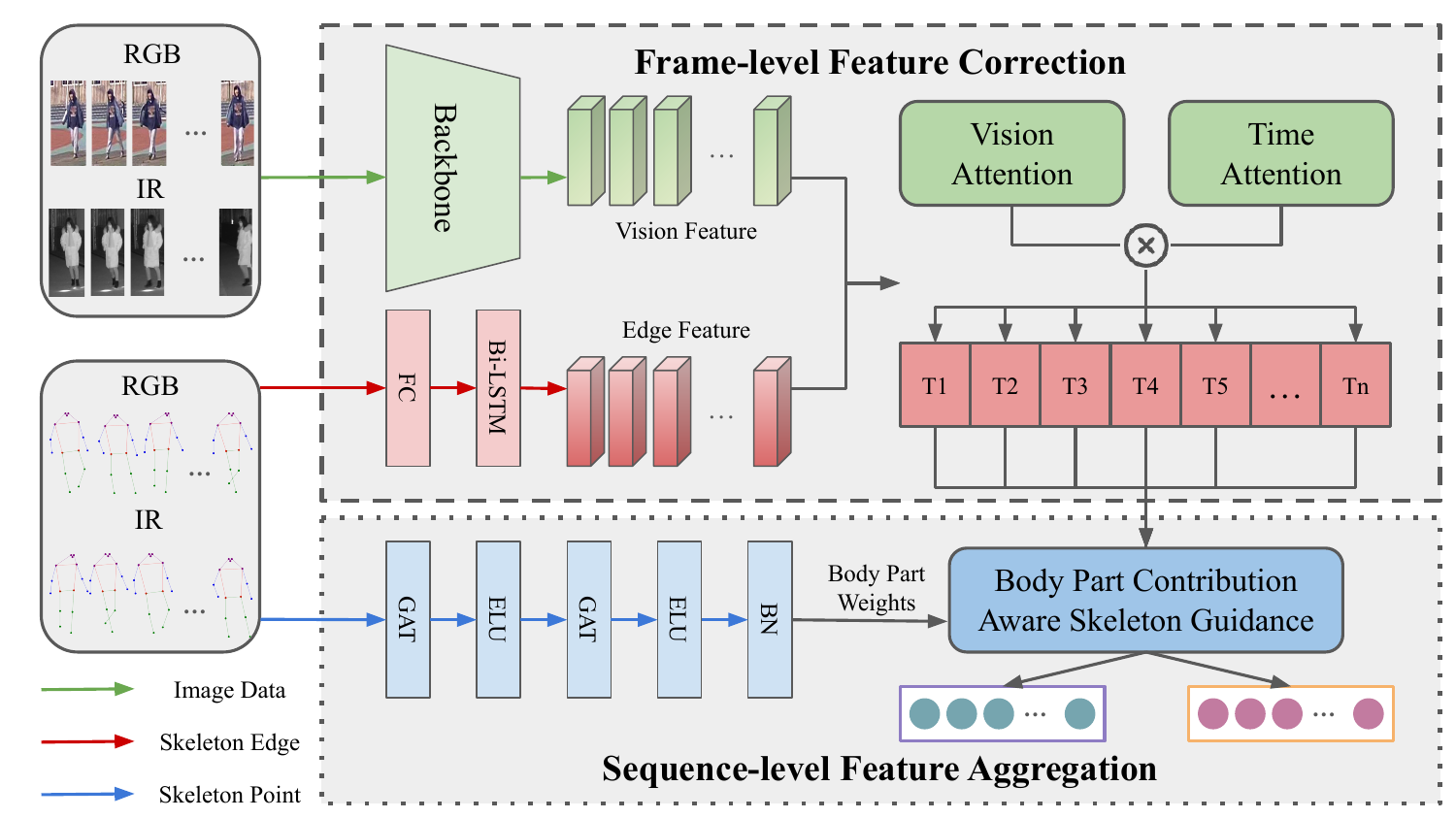} 
\caption{Overview of the STAR method for VVI-ReID. The final output represents the identity probabilities of pedestrians in different modalities. At the frame level, skeleton connections refine individual frame features, and at the sequence level, a skeleton graph sequence aggregates features to enhance global accuracy. STAR improves spatial-temporal feature extraction in both visible and infrared modalities.}
\label{main-dataflow}
\end{figure*}

\subsection{Skeleton-based Person ReID}

In recent years, there have been attempts to apply skeleton-based methods to the field of person re-identification \cite{zhou2023learning,cormier2024enhancing,ren2024survey}. Existing skeleton-based approaches have been applied in both cross-modal and single-modal image-based person re-identification. Some studies focus on single-modal image person re-identification. For instance, HSMLP-Reid \cite{wang2020human} combines a novel partial pedestrian segmentation method with global skeleton information to address the challenges posed by background and local pose variations. Other studies have investigated the application of skeletons in cross-modal image-based person re-identification. For instance, \cite{miao2023exploring} employed pose estimation as a supplementary learning component to enhance VI-ReID within an end-to-end framework. These studies have achieved significant success in the field of image-based person re-identification. 

However, due to the challenges of low-quality and occluded videos, these skeleton-based methods cannot be directly applied to the domain of VVI-ReID. Our research is the first to introduce skeleton-based methods to VVI-ReID, using skeleton modality to temporally guide the video information. We utilize skeleton joints for sequence-level guidance to capture fine-grained temporal information across video frames, and skeleton edges for frame-level guidance to refine the visual features of individual frames. This two-level approach enhances the extraction of spatial-temporal features, improving accuracy in both visible and infrared modalities.
\section{Method}

In the STAR method, we introduce skeleton features as guidance to address the challenges of spatial-temporal feature extraction in VVI-ReID caused by modality discrepancies, low infrared video quality, and partial occlusions. Our design focuses on: (a) leveraging the structured information of skeletons to refine visual features and mitigate the effects of noise and occlusions; and (b) utilizing skeleton features to optimize cross-frame feature aggregation, ensuring global consistency and accuracy of spatial-temporal features. Figure \ref{main-dataflow} clearly depicts how skeleton information enhances feature learning for VVI-ReID at different levels.

\subsection{Skeleton Guided Frame-level Correction}

To enhance visual feature extraction in person re-identification tasks under complex scenarios, we integrate human skeleton information as structured guiding signals\cite{hrnet}. This approach addresses challenges such as low image quality, varying lighting conditions, and partial occlusions, which often degrade the performance of traditional visual feature-based methods. 

\textbf{Skeleton Edge Feature Representation.}  
Skeletal features inherently capture the structured and stable nature of human motion, such as keypoint positions and joint connection lengths. These features exhibit robustness to lighting variations and partial occlusions, making them ideal for compensating deficiencies in visual features. 

To represent these features, we compute the pairwise spatial relationships between skeleton joints. Let $x_i$ and $y_i$ represent the coordinates of joint $i$, and $x_j$ and $y_j$ represent the coordinates of joint $j$ \cite{COCO}. The pairwise distance $d_{ij}(t)$ and relative angle $\theta_{ij}(t)$ between joints $i$ and $j$ at time step $t$ are defined as:
\begin{equation}
    d_{ij}(t) = \sqrt{(x_j - x_i)^2 + (y_j - y_i)^2},
\end{equation}
\begin{equation}
    \theta_{ij}(t) = \tan^{-1}\left(\frac{y_j - y_i}{x_j - x_i}\right).
\end{equation}

These pairwise features ${G}_{ij}(t)$ are processed through a fully connected (FC) layer and modeled temporally using a long short-term memory (LSTM) network to extract the temporal skeleton characteristics ${T}_t$:
\begin{equation}
    {T}_t = \mathrm{LSTM}(\{\mathrm{FC}({G}_{ij}(t))\}_{t=1}^{T}),
\end{equation}
where ${T}_t$ represents the temporal guidance derived from skeleton features at time step $t$.

\textbf{Time Attention Mechanism.}  
To refine frame-level feature extraction, we employ a time attention mechanism to establish temporal dependencies between visual features ${X}_t$ and skeleton-guided temporal features ${T}_t$. Specifically, a cross-attention operation\cite{DBLP:conf/nips/VaswaniSPUJGKP17} enhances the visual features ${H}_{t,t}$ by leveraging the temporal skeleton guidance:
\begin{align}
    {H}_{t,t} &= \mathrm{CrossAttn}({X}_t, {T}_t, {T}_t), \\
    {W}_t     &= \mathrm{SE}({H}_{t,t}),
\end{align}
where $\mathrm{CrossAttn}(\cdot)$ denotes a cross-attention mechanism, and $\mathrm{SE}(\cdot)$ represents a squeeze-and-excitation layer\cite{SEmodule} that generates adaptive weights ${W}_t$ for enhanced temporal features.

\textbf{Visual Attention Mechanism.}  
To further integrate visual and skeleton features, a visual attention mechanism enhances the visual features ${H}_{o,t}$ using skeleton-guided attention:
\begin{equation}
    {H}_{o,t} = \mathrm{CrossAttn}({T}_t, {X}_t, {X}_t).
\end{equation}

The refined visual features ${Y}_t$ are obtained by combining the outputs of the visual and temporal attention mechanisms. These features are processed through a feed-forward network (FFN)\cite{FFN} and combined with the adaptive weights from the SE layer, along with residual connections for stability:
\begin{equation}
    {Y}_t = \mathrm{FFN}([{H}_{o,t}; {H}_{t,t}]) \otimes {W}_t + {X}_t + \epsilon,
\end{equation}
where $[\cdot]$ denotes feature concatenation, $\otimes$ represents element-wise multiplication, and $\epsilon$ is a small constant (e.g., $10^{-3}$) to enhance numerical stability.

This comprehensive framework for skeleton-guided frame-level correction effectively addresses the limitations of visual feature extraction, ensuring robustness in complex scenarios and providing a stable foundation for subsequent temporal modeling.

\subsection{Sequence-level Feature Aggregation}
In video sequences, directly applying conventional pooling methods to long-sequence features often fails to capture spatial-temporal consistency\cite{SAADG,ye2021deep}, mitigate modality discrepancies, or preserve discriminative features. To address these limitations, we propose a skeleton-mediated feature aggregation mechanism that leverages the spatial-temporal information encoded in skeleton trajectories. This mechanism models the contributions of different body parts to spatial-temporal features, thereby enhancing global feature representations.

\textbf{Body Part Contribution Aware Skeleton Guidance.} Skeletons encapsulate unique patterns of individual motion. To exploit these patterns, we construct a spatial-temporal graph \( G = (\mathcal{V}, \mathcal{E}) \), where \( \mathcal{V} \) represents key points across frames, and \( \mathcal{E} \) captures both spatial and temporal relationships:

\begin{itemize}
    \item \textbf{Spatial Connection}: For a frame at time \( t \), skeletal key points are connected based on their structural relationships:
    \begin{equation}
        \mathcal{E}_{\text{spatial}}^{(t)} = \left\{ \left( v_i^{(t)}, v_j^{(t)} \right) \mid v_i, v_j \in \mathcal{V};\ (i, j) \in \mathcal{E}_S \right\},
    \end{equation}
    where \( \mathcal{E}_S \) defines the anatomical connections between key points.
    \item \textbf{Temporal Connection}: To model motion dynamics, corresponding key points in consecutive frames are connected:
    \begin{equation}
        \mathcal{E}_{\text{temporal}} = \left\{ \left( v_i^{(t)}, v_i^{(t+1)} \right) \mid v_i \in \mathcal{V};\ t = 1, \dots, T-1 \right\}.
    \end{equation}
\end{itemize}

\begin{figure}[t]
\centering
\includegraphics[width=0.9\columnwidth]{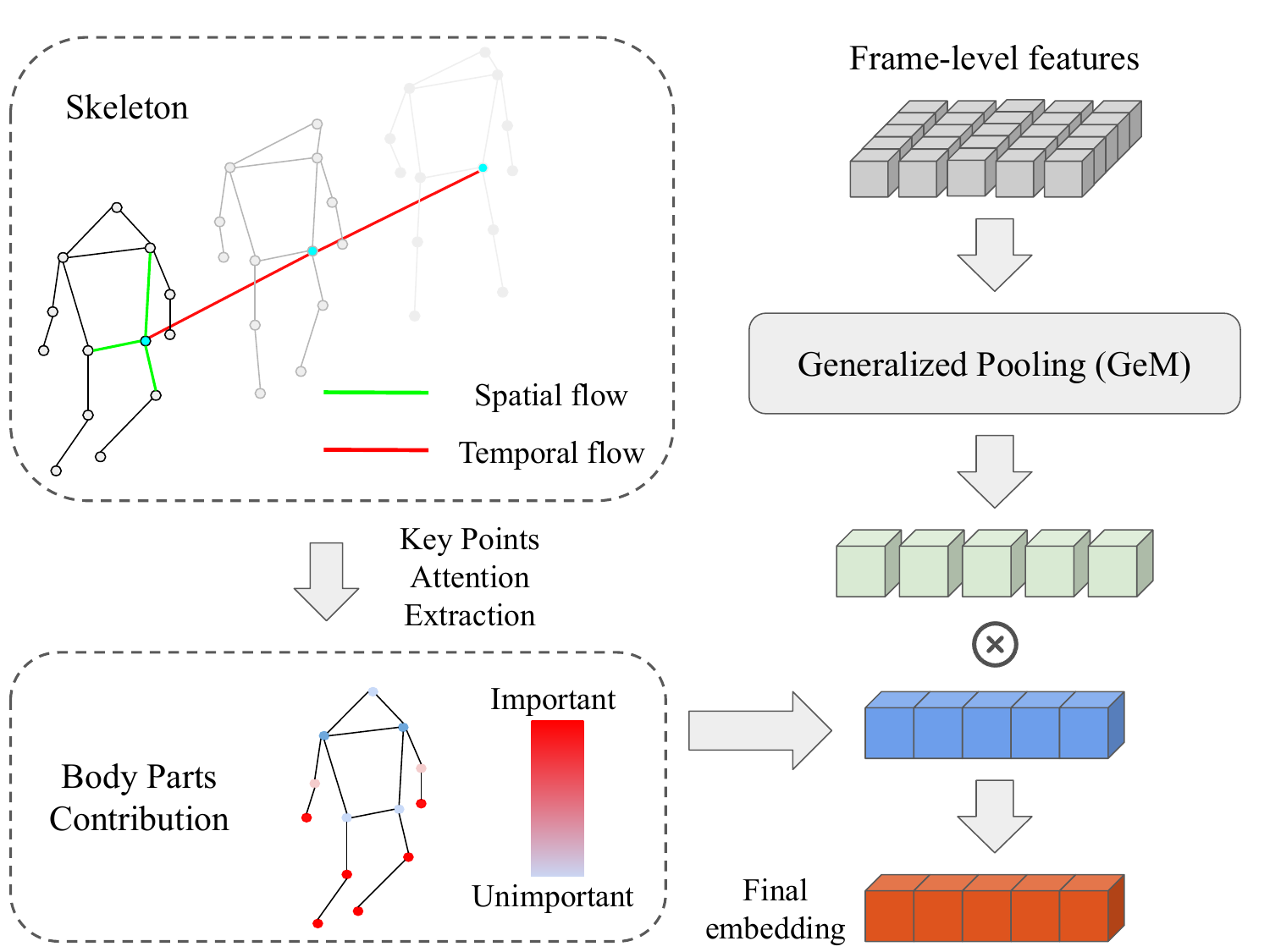} 
\caption{\textbf{Body Part Contribution Aware Skeleton Guidance}: Illustration of the spatial-temporal skeleton graph and the dynamic pooling mechanism that selectively integrates global features based on Body Part Contributions.}
\label{BAGP}
\end{figure}

The unified edge set \( \mathcal{E} = \bigcup_{t=1}^T \mathcal{E}_{\text{spatial}}^{(t)} \cup \mathcal{E}_{\text{temporal}} \) enables the graph to capture spatial configurations and temporal transitions, as shown in Figure \ref{BAGP}.

To enhance the modeling capacity of the skeleton graph, we employ a Graph Attention Network (GAT)\cite{GAT}. The adaptive attention mechanism of GAT dynamically learns importance weights for different key points and edges, allowing the model to prioritize body parts based on their contextual significance. For example, during person motion, the dynamics of the legs may contribute more prominently to spatial-temporal features than those of the arms. GAT effectively identifies and incorporates such variations.

The attention mechanism assigns a weight \( \alpha_{ij} \) to each edge \( (i, j) \), reflecting the relative importance of the connection between node \( i \) and node \( j \). These weights are normalized across all neighbors \( \mathcal{N}(i) \) of node \( i \) to ensure they form a valid probability distribution. Using the learned attention weights \( \alpha_{ij} \), the features of each node are updated by aggregating information from its neighbors as follows:
\begin{equation}
    \mathbf{h}_i' = \mathrm{ELU}\left(\sum_{j \in \mathcal{N}(i)} \alpha_{ij} \mathbf{W}\mathbf{h}_j\right),
\end{equation}
where \( \mathbf{h}_i' \) is the updated feature of node \( i \), \( \mathbf{W} \) is a learnable weight matrix, and \(\mathrm{ELU}(\cdot)\) is a non-linear activation function\cite{ELU}. This mechanism allows the model to focus on the most relevant graph connections while updating node features.

We stack two GAT layers, applying an ELU activation after each to enhance non-linearity and capture complex relationships. A Batch Normalization (BN) layer follows the second GAT layer to stabilize feature distributions, as shown in Figure \ref{main-dataflow}.

\textbf{Sequence-level Feature Aggregation.} Building on skeleton-guided attention weights, we design a generalized pooling strategy that integrates contributions from different body parts into spatial-temporal features. This strategy is grounded in the parameterized Generalized Mean (GeM) pooling\cite{GeM}, which unifies max and average pooling through an adjustable parameter \( p \):
\begin{equation}
    {g}(t) = \left( \frac{1}{N} \sum_{i=1}^N |{f}_i(t)|^p \right)^{\frac{1}{p}},
\end{equation}
where \( N \) is the number of features at time \( t \), and \( {f}_i(t) \) represents individual feature responses. 

The skeleton-guided attention weights are aggregated using the outputs of the GAT as follows:
\begin{equation}
    \alpha_p(n) = \frac{\sum_{i \in \mathcal{B}(n)} \alpha_{ij} \mathbf{h}_j}{|\mathcal{B}(n)|},
\end{equation}
where $\mathcal{B}(n)$ represents the set of nodes corresponding to body part \( n \), and $\alpha_{ij}$ is the GAT-derived attention coefficient. The aggregated feature vector \( {G} \) for the sequence is then computed as:
\begin{equation}
    {G} = \frac{1}{T} \sum_{t=1}^T \sum_{n} \alpha_p(n) {g}(t),
\label{SLM}
\end{equation}
where \( \alpha_p(n) \) represents the aggregated skeleton-guided attention weights for body part \( n \), and \( T \) denotes the total number of frames, as shown in Figure \ref{BAGP}.

This dynamic weighting strategy ensures that the contributions of different body parts are adaptively enhanced by the global feature representation. By leveraging body part awareness attention, our model dynamically emphasizes key body parts, effectively capturing spatial-temporal variations in sequence features. This approach strengthens the robustness and discriminative power of global representations, improving performance in pedestrian retrieval tasks.

\subsection{Training Objectives}

To further enhance the integration of skeletal data with feature learning, we introduce a skeleton consistency loss \(L_{KL}\) to \textbf{align the feature spaces} of joint information and edge information. Specifically, we constrain the features by applying a bidirectional KL divergence between the output of the last time step of the LSTM (used for edge information processing) and the weighted joint features after passing through a fully connected layer.The bidirectional skeleton consistency loss \(L_{KL}\) is formulated as:
\begin{equation}
    L_{KL} = \frac{1}{2} \sum_{t=1}^{T} \sum_{i} \left[ \mathbf{P}_i(t) \log \frac{\mathbf{P}_i(t)}{\mathbf{Q}_i(t)} + \mathbf{Q}_i(t) \log \frac{\mathbf{Q}_i(t)}{\mathbf{P}_i(t)} \right],
\end{equation}
where \(\mathbf{P}(t)\) represents the probability distribution obtained from the output of the last LSTM time step, and \(\mathbf{Q}(t)\) represents the weighted joint features after processing through the fully connected layer. \(\mathbf{P}_i(t)\) and \(\mathbf{Q}_i(t)\) are the corresponding probabilities for each feature dimension \(i\) at time step \(t\).

The total objective function of our STAR method is defined as:
\begin{equation}
    L = L_{id} + L_{tri} + L_{KL} + L_{SAD},
\end{equation}
where \(L_{SAD}\) is the data augmentation loss in the article \cite{SAADG}, \(L_{id}\) and \(L_{tri}\) are the identity loss and triplet loss based on different level representations guided temporally after the integration.

This comprehensive objective function ensures that the integration of skeletal data is effectively guided across temporal representations, promoting robust feature alignment and enhancing the overall performance of the STAR method.

\section{Experiments}
\subsection{Experimental Settings}
\begin{table*}[t]\small
\centering
\begin{tblr}{
  width=\textwidth,
  column{4-8,9-14} = {c},
  column{2-3} = {c},
  cell{1}{1} = {r=2}{c},
  cell{1}{2} = {r=2}{c},
  cell{1}{3} = {r=2}{c},
  cell{1}{4} = {c=5}{c},
  cell{1}{9} = {c=5}{c},
  vline{2,3,4,9} = {1-14}{},
  hline{1,3,9,13,14} = {-}{},
  hline{2} = {1-14}{},
}
Methods & Sources  & Type  & Infrared to Visible (I2V) &    &    &    &     & Visible to Infrared (V2I) &    &    &    &     \\
       &          &       & rank-1                        & rank-5 & rank-10 & rank-20 & mAP & rank-1                        & rank-5 & rank-10 & rank-20 & mAP \\
LbA    & ICCV’21  & Image & 46.38                     & 65.29 & 72.23 & 79.41 & 30.69 & 49.30                     & 69.27 & 75.90 & 82.21 & 32.38 \\
MPANet & CVPR’21  & Image & 46.51                     & 63.07 & 70.51 & 77.77 & 35.26 & 50.32                     & 67.31 & 73.56 & 79.66 & 37.80 \\
DDAG   & ECCV’20  & Image & 54.62                     & 69.79 & 76.05 & 81.50 & 39.26 & 59.03                     & 74.64 & 79.53 & 84.04 & 41.50 \\
VCD    & CVPR’21  & Image & 54.53                     & 70.01 & 76.28 & 82.01 & 41.18 & 57.52                     & 73.66 & 79.38 & 83.61 & 43.45 \\
CAJL   & ICCV’21  & Image & 56.59                     & 73.49 & 79.52 & 84.05 & 41.49 & 60.13                     & 74.62 & 79.86 & 84.53 & 42.81 \\
SGIEL  & CVPR’23  & Image & 67.65                     & 80.32 & 84.73 & -     & 52.30 & 70.23                     & 82.19 & 86.11 & -     & 52.54 \\
MITML  & CVPR’22  & Video & 63.74                     & 76.88 & 81.72 & 86.28 & 45.31 & 64.54                     & 78.96 & 82.98 & 87.10 & 47.69 \\
IBAN   & TCSVT’23 & Video & 65.03                     & 78.34 & 82.98 & 87.19 & 48.77 & 69.58                     & 81.51 & 85.43 & 88.78 & 50.96 \\
SAADG  & MM'23   & Video & 69.22                     & 80.61 & 85.03 & 88.66 & 53.77 & 73.13                     & 83.47 & 86.86 & 89.72 & 56.09 \\
CST    & TMM'24 & Video & 69.44                     & 81.13 & \textbf{85.75} & \textbf{89.70} & 51.16 & 72.64                     & 83.41 & 86.74 & 89.78 & 53.00 \\
\textbf{Ours}   &          & \textbf{Video} & \textbf{69.57}                    & \textbf{81.54} & 84.58  & 88.97  & \textbf{56.01} & \textbf{73.25}                   & \textbf{84.26} & \textbf{87.92}  & \textbf{91.04}  & \textbf{58.56} 
\end{tblr}
\caption{Comparisons of STAR with SOTA methods on HITSZ-VCM in terms of CMC(\%) and mAP(\%).}
\label{Compare-SOTA}
\end{table*}

\subsubsection{Datasets.}
In this study, experiments were conducted on the HITSZ-VCM dataset\cite{MITML}, a unique video-based visible-infrared person re-identification dataset. Each individual's track is represented by a sequence of 24 continuous frames. The dataset consists of 251,452 RGB images and 211,807 IR images, capturing 927 identities from 12 non-overlapping cameras. These are divided into 11,785 RGB tracks and 10,078 IR tracks. Following the same methodology as \cite{MITML}, we partitioned the HITSZ-VCM dataset into training and testing sets. The training set comprises 232,496 images and 11,061 trajectories, covering 500 identities, while the testing set includes 230,763 images and 10,802 trajectories, covering 427 identities.

\subsubsection{Evaluation Protocols.} 
The performance of STAR is evaluated using Rank-k accuracy and mAP (mean Average Precision), both of which are widely employed in cross-modality person re-identification.

\subsubsection{Implementation Details.}
We used the SAD module from SAADG\cite{SAADG} as our backbone network and constructed the model mainly using MindSpore on 8 $\times$ Ascend 910. We set the batch size to 8, with 2 instances per identity in each modality. During the model optimization process, we used an SGD optimizer with an initial learning rate of 0.1 and a weight decay of $5\times 10^{-4}$. The initial learning rate is set at 0.1, employing a linear warmup strategy for the first 10 epochs. Subsequently, it is reduced by a factor of 0.1 at the 60th and 100th epochs. We set the backbone network's learning rate to one-tenth of the global rate to improve convergence through transfer learning. Before training, all images are resized to 288 × 144 pixels. This is followed by padding each image by 10 pixels on all sides and then performing a random crop. 

\subsection{Comparison with State-of-the-art Methods}

In this subsection, we conducted a comparative analysis of our STAR approach with state-of-the-art methods, encompassing both image-based and video-based approaches. Specifically, for image-based methods, we considered the following approaches: LbA \cite{LbA}, MPANet \cite{MPANet}, DDAG \cite{DDAG}, VCD \cite{VCD}, CAJL \cite{CAJL}, and SGIEL \cite{SGIEL}. For video-based methods, we included MITML \cite{MITML}, IBAN \cite{IBAN}, SAADG \cite{SAADG}, and CST \cite{feng2024cross}.

As illustrated in Table \ref{Compare-SOTA}, our proposed method demonstrates significant effectiveness over state-of-the-art approaches in both infrared-visible (I2V) and visible-infrared (V2I) modes. We achieve a Rank-1 accuracy of 69.57\% in the I2V scenario and 73.25\% in the V2I scenario, surpassing recent methods like SAADG and CST. Additionally, our method demonstrates a significant improvement in mAP, with 56.01\% for I2V and 58.56\% for V2I. These gains are attributed to the STAR method we proposed, which enhances both frame-level and sequence-level feature extraction, ensuring more accurate and robust identification under challenging conditions.

\subsection{Ablation Study}

In this section, we conducted extensive experiments on the HITSZ-VCM dataset in visible-to-infrared (V2I) and infrared-to-visible (I2V) modes. Through ablation studies, we validated the contribution of pedestrian pose data within STAR to VVI-ReID and assessed the effectiveness of each module within the network. The SAADG backbone was used as the baseline for this experiment.
\begin{table}
\centering
\begin{tblr}{
  row{even} = {c},
  row{1,2,3} = {c},
  row{5} = {c},
  cell{1}{1} = {c=2}{},
  cell{1}{3} = {c=2}{},
  cell{1}{5} = {c=2}{},
  vline{3,5} = {1,2,3-6}{},
  hline{1-3,7} = {-}{},
}
Extraction Level &      & I2V    &       & V2I    &       \\
Frame    & Sequences & rank-1 & mAP   & rank-1 & mAP   \\
\ding{55}     &  \ding{55}    & 68.36  & 54.97 & 70.83  & 55.89 \\
\ding{55}      & \ding{51}   & 68.71  & 55.93 & 71.01  & 56.16 \\
\ding{51}     & \ding{55}   & 68.88  & 55.04 & 72.61  & 57.92 \\
\ding{51}     & \ding{51}  & \textbf{69.57}  & \textbf{56.01} & \textbf{73.25}  & \textbf{58.56}

\end{tblr}
\caption{Performance Evaluation of Frame-level and Sequence-level Skeleton Information Extraction in the STAR Method. }
\label{SmallTimeGuidance}
\end{table}

\textbf{Effectiveness of Edge Information Extraction.}  
As shown in Table \ref{SmallTimeGuidance}, incorporating edge information extraction significantly improves performance. Compared to the network without this component, the network with edge information achieves a 0.64\% improvement in Rank-1 accuracy and a 0.65\% increase in mAP. This design effectively minimizes the adverse effects of occlusion and viewpoint changes on image feature extraction.

\textbf{Effectiveness of Key-point Information Extraction.}  
As shown in Table \ref{SmallTimeGuidance}, removing the key-point information extraction module led to a significant decline in performance. Including this module resulted in a 2.24\% improvement in Rank-1 accuracy and a 2.4\% increase in mAP compared to the network without it. This demonstrates the importance of leveraging key-point information to ensure the consistency of sequence-level temporal features.

\begin{table*}
\centering
\begin{tblr}{
  row{1,2,3,4,5,6} = {c},
  cell{1}{1} = {r=2}{},
  cell{1}{2} = {c=5}{},
  cell{1}{7} = {c=5}{},
  vline{2} = {1-2}{},
  vline{7} = {2}{},
  vline{2,7} = {1-7}{},
  hline{1,3,7} = {-}{},
  hline{2} = {2-11}{},
  row{6} = {font=\bfseries},
}
Methods       & Infrared to Visible (I2V) &        &         &         &       & Visible to Infrared (V2I) &        &         &         &       \\
              & rank-1                    & rank-5 & rank-10 & rank-20 & mAP   & rank-1                    & rank-5 & rank-10 & rank-20 & mAP   \\
Baseline      & 66.76                    & 80.24 & 83.51  & 87.55  & 54.05 & 70.45                   & 82.06 & 86.23  & 89.19  & 55.45 \\
Baseline + FG  & 68.53                    & 80.28 & 83.99  & 87.81  & 55.01 & 71.01                   & 83.77 & 86.20  & 89.88  & 56.16 \\
Baseline + SG & 69.04                    & 80.29 & 84.58  & 87.93  & 55.91 & 73.15                   & 84.10 & 86.92  & 90.97  & 58.22 \\
Full Method   & 69.57                    & 81.54 & 84.58  & 88.97  & 56.01 & 73.25                   & 84.26 & 87.92  & 91.04  & 58.56
\end{tblr}
\caption{Performance evaluation of Frame-level Guidance(FG) and Sequence-level Guidance(SG). Experiments based on the MindSpore.}
\label{TimeGuidance}
\end{table*}

\textbf{Effectiveness of Different Level Guidance.}  
To validate the impact of frame-level and sequence-level guidance on the experimental results, we conducted experiments on the HITSZ-VCM dataset. As shown in Table \ref{TimeGuidance}, compared to the baseline, incorporating frame-level guidance provided a gain of 0.56\% in Rank-1 accuracy and 0.71\% in mAP. Sequence-level guidance further improved Rank-1 accuracy and mAP by 2.70\% and 2.77\%, respectively, clearly demonstrating its ability to effectively capture temporal features from videos. More importantly, when both frame-level and sequence-level guidance are combined, performance improves further, with Rank-1 accuracy and mAP increasing by 2.80\% and 3.11\% over the baseline. This indicates that frame-level and sequence-level guidance complement each other, enabling the model to capture richer spatial-temporal features.

\subsection{Visualization Analysis}
\begin{figure}[ht]
\centering
\includegraphics[width=0.8\columnwidth]{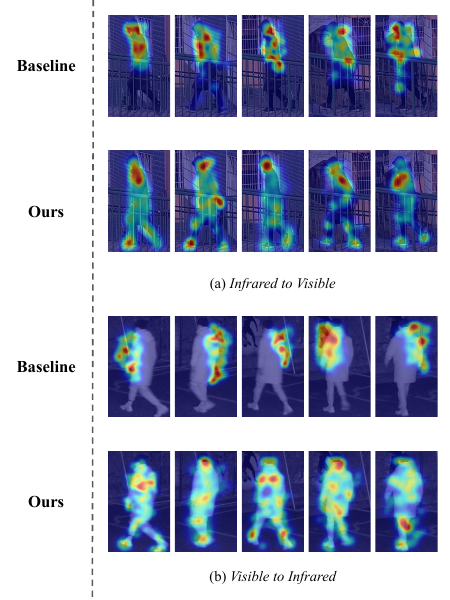} 
\caption{Visualization of heat maps by using Grad-CAM \cite{Selvaraju_2017_ICCV}, comparing the baseline and STAR, showing that STAR better highlights discriminative regions.}
\label{vis}
\end{figure}

To demonstrate the effectiveness of our STAR approach, we randomly sampled video clips to visualize retrieval results using heatmaps in both modes. Additionally, we compared the effectiveness of the STAR method with the baseline through these visualizations. Our approach successfully captures the key features of pedestrians, even under challenging conditions such as partial occlusions. As illustrated in Figure \ref{vis}, our method highlights key skeletal regions, including arms, knees, and legs, resulting in a more accurate representation of the features.

\subsection{Impact of Sequence Length on Performance}
\begin{figure}
    \centering
    \hspace*{0cm} 
    \includegraphics[width=0.7\linewidth]{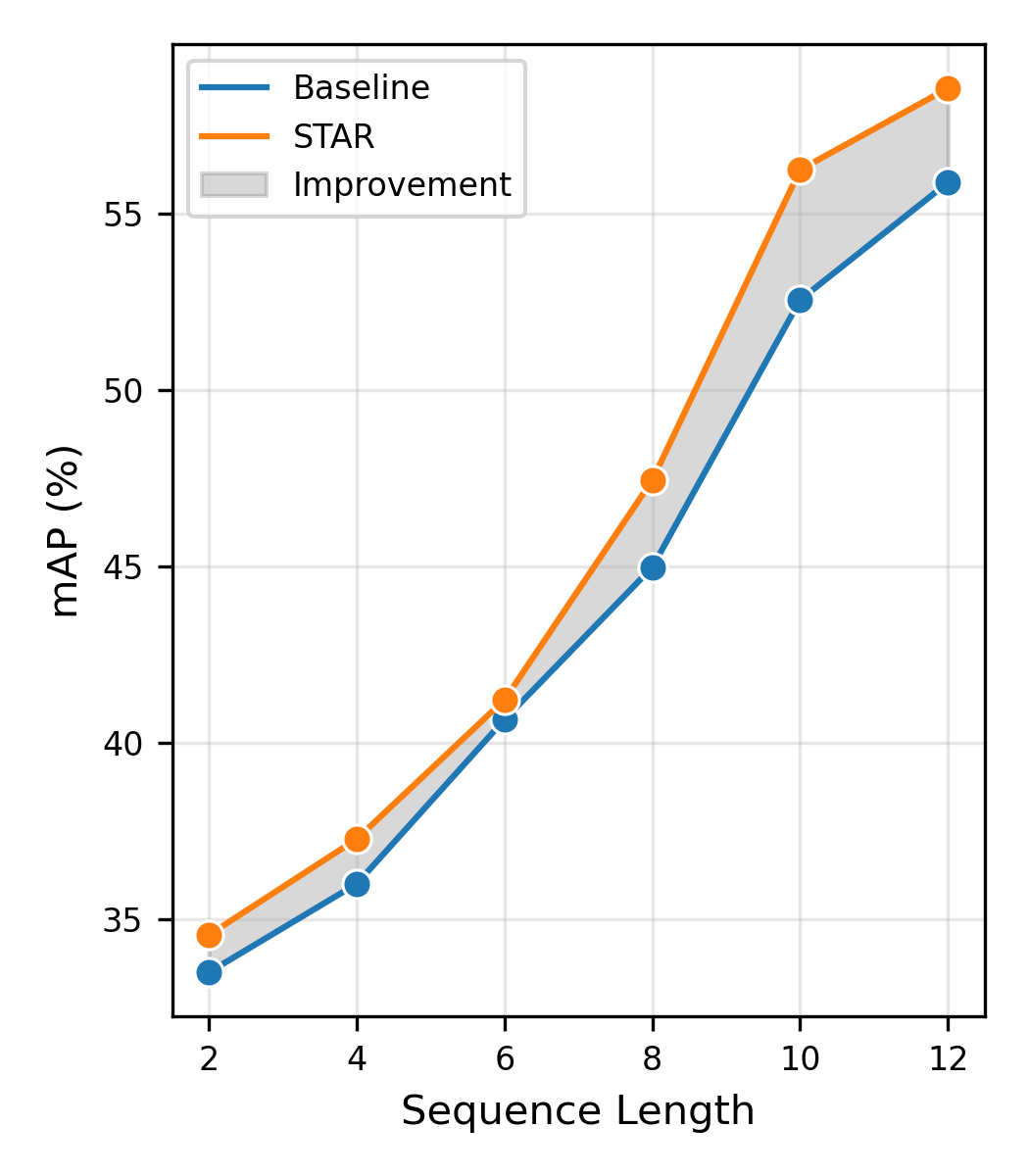}
    \caption{Comparison of mAP (\%) at different sequence lengths for the baseline model and the STAR model. The shaded area represents the improvement margin achieved by the STAR model over the baseline. This comparison results were obtained using the MindSpore framework.}
    \label{fig:seq_map}
\end{figure}

Sequence length plays a crucial role in video-based person re-identification, significantly influencing retrieval performance. To evaluate the effectiveness of skeletal features in refining spatial-temporal structures and enhancing feature aggregation, we conducted a series of experiments with varying sequence lengths. The results of these experiments are shown in Figure~\ref{fig:seq_map}.

Figure~\ref{fig:seq_map} illustrates that the STAR model consistently outperforms the baseline model across all sequence lengths. Notably, in long-sequence scenarios (e.g., sequence lengths of 10 to 12), the STAR model achieves a substantial performance boost, as highlighted by the shaded improvement region. This indicates the STAR model's ability to capture critical motion and posture information, effectively compensating for the limited temporal context. By refining the spatial-temporal structure and aggregating skeletal features, the STAR model demonstrates its advantage in scenarios with restricted temporal information. These results emphasize the STAR model's robustness and adaptability across varying sequence lengths, particularly in scenarios where temporal information is constrained.

\subsection{Hyperparameter Analysis}
In designing the sequence-level guidance for the skeleton data, we predefined multiple values of the hyperparameter \( p \) to generate features of varying granularity. According to Equation (3), the value of \( p \) determines the effect of feature integration: when \( p = 1 \), the resulting feature corresponds to the average grouping of frame-level features; when \( p \to \infty \), it degenerates to max pooling; and when \( p < 1 \), it emphasizes smaller feature values.

The objective of this design is to provide candidates with multiscale features, enabling the aggregation of contributions from different body parts. To achieve this, we selected \( p = 1 \) as a critical threshold, sampling several values both below (\( p < 1 \)) and above (\( p > 1 \)) this point to analyze their impact on the representation of the global feature space. 

\begin{figure}[ht]
\centering
\includegraphics[width=0.9\columnwidth]{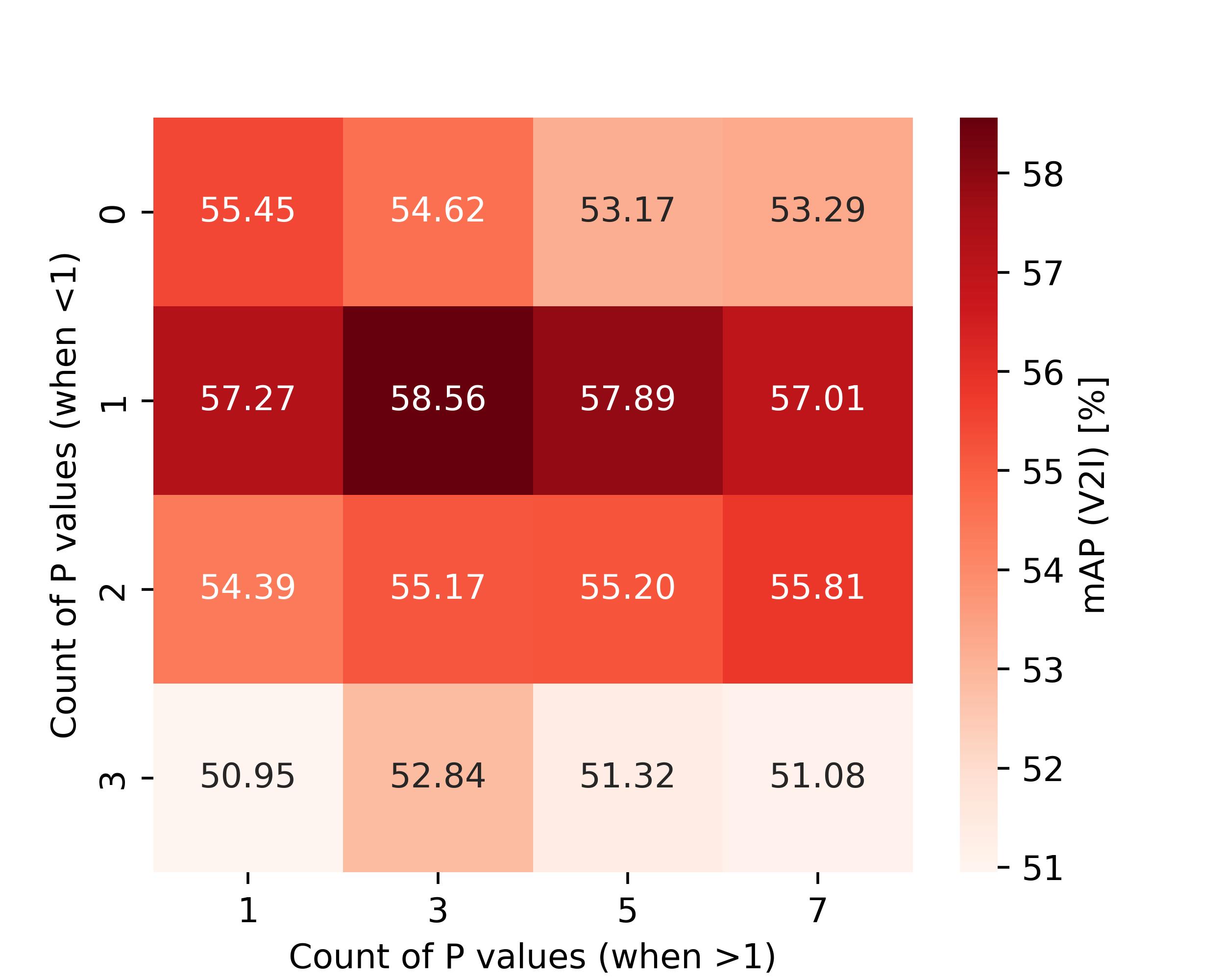} 
\caption{Impact of the hyperparameter \( p \) combination on global feature representation. The two axes in the heatmap below represent the number of \( p \) values chosen below and above the threshold, respectively.}

\label{heatmap}
\end{figure}

As shown in Figure~\ref{heatmap}, using multiple \( p < 1 \) values results in a significant performance drop. The optimal configuration includes one \( p < 1 \) value and three \( p > 1 \) values. This combination strikes a balance, as \( p < 1 \) pooling effectively captures subtle variations in features, while overusing it increases sensitivity to noise, leading to model instability.
\section{Conclusion}
In this paper, we propose a skeleton-guided spatial-temporal feature learning method for Video-based visible-infrared person re-identification (VVI-ReID). This method effectively addresses the challenge of spatial-temporal information extraction in videos, particularly in the context of infrared videos, by utilizing a two-level skeleton-guided strategy: frame-level and sequence-level. At the frame level, the robust structured skeleton information refines the visual features of individual frames, improving feature accuracy. At the sequence level, we designed a body part contribution aware skeleton guidance module to learn the contributions of different body parts, further enhancing the accuracy of global features. 
Extensive experiments conducted on benchmark datasets demonstrate the advanced capabilities and effectiveness of our method in VVI-ReID task. Ablation studies further validate the importance and contribution of each module. 

\textbf{Acknowledgments.} Thanks for the support provided by MindSpore Community. Part of the experimental data presented in this paper was obtained using the MindSpore framework.
{
    \small
    \bibliographystyle{ieeenat_fullname}
    \bibliography{main}
}


\end{document}